\documentclass[sigconf]{acmart}
\AtBeginDocument{%
  }
\usepackage[table]{xcolor} 
\usepackage{booktabs}      
\usepackage{amsmath}       
\usepackage{tabularx}     
\usepackage{graphicx}
\usepackage{enumitem}
\usepackage{multirow}
\usepackage{array}
\usepackage{enumitem}

\usepackage{csvsimple}
\usepackage{subcaption}
\usepackage{xfp} 
\usepackage{pgf}

\newcommand{\hmax}[2]{%
    \ifnum\fpeval{#1 == #2}=1 \textbf{#1}\else #1\fi%
}
\newcommand{\hmin}[2]{%
    \ifnum\fpeval{#1 == #2}=1 \textbf{#1}\else #1\fi%
}
\acmDOI{XXXXXXX.XXXXXXX}

\copyrightyear{2026}
\acmYear{2026}
\setcopyright{acmlicensed} 
\acmConference[Conference acronym ’XX]{Make sure to enter the correct conference title from your rights confirmation email}
\acmBooktitle{}

\newcommand{\ModelName}{MobilityBench}

\begin{document}

\title{MobilityBench: A Benchmark for Evaluating Route-Planning Agents in Real-World Mobility Scenarios}
\author{Zhiheng Song}
\authornote{Both are co-first authors and contribute equally to this work.}
\affiliation{%
  \institution{Computer Network Information Center, Chinese Academy of Sciences\\ AMAP, Alibaba Group}
  \city{Beijing}
  \country{China}}
\email{songzhiheng0426@gmail.com}

\author{Jingshuai Zhang}
\authornotemark[1]
\affiliation{%
  \institution{AMAP, Alibaba Group}
  \city{Beijing}
  \country{China}}
\email{zhangjingshuai0@gmail.com}

\author{Chuan Qin}
\authornote{Corresponding authors.}
\affiliation{%
  \institution{Computer Network Information Center, Chinese Academy of Sciences}
  \city{Beijing}
  \country{China}}
\email{chuanqin0426@gmail.com}

\author{Chao Wang}
\affiliation{%
  \institution{Independent Researcher}
  \country{China}}
\email{chadwang2012@gmail.com}

\author{Chao Chen}
\affiliation{%
  \institution{AMAP, Alibaba Group}
  \city{Beijing}
  \country{China}}
\email{cc201598@alibaba-inc.com}

\author{Longfei Xu}
\affiliation{%
  \institution{AMAP, Alibaba Group}
  \city{Beijing}
  \country{China}}
\email{longfei.xl@alibaba-inc.com}

\author{Kaikui Liu}
\affiliation{%
  \institution{AMAP, Alibaba Group}
  \city{Beijing}
  \country{China}}
\email{damon@alibaba-inc.com}

\author{Xiangxiang Chu}
\affiliation{%
  \institution{AMAP, Alibaba Group}
  \city{Beijing}
  \country{China}}
\email{cxxgtxy@gmail.com}

\author{Hengshu Zhu}
\authornotemark[2]
\affiliation{%
  \institution{Computer Network Information Center, Chinese Academy of Sciences}
  \city{Beijing}
  \country{China}}
\email{zhuhengshu@gmail.com}

\renewcommand{\shortauthors}{Song \& Zhang et al.}

\begin{abstract}
Route-planning agents powered by large language models (LLMs) have emerged as a promising paradigm for supporting everyday human mobility through natural language interaction and tool-mediated decision making. However, systematic evaluation in real-world mobility settings is hindered by diverse routing demands, non-deterministic mapping services, and limited reproducibility. In this study, we introduce \textbf{\ModelName}, a scalable benchmark for evaluating LLM-based route-planning agents in real-world mobility scenarios. \ModelName\ is constructed from large-scale, anonymized real user queries collected from Amap and covers a broad spectrum of route-planning intents across multiple cities worldwide. To enable reproducible, end-to-end evaluation, we design a deterministic API-replay sandbox that eliminates environmental variance from live services. We further propose a multi-dimensional evaluation protocol centered on outcome validity, complemented by assessments of instruction understanding, planning, tool use, and efficiency. Using \ModelName, we evaluate multiple LLM-based route-planning agents across diverse real-world mobility scenarios and provide an in-depth analysis of their behaviors and performance.
Our findings reveal that current models perform competently on Basic information retrieval and Route Planning tasks, yet struggle considerably with Preference-Constrained Route Planning, underscoring significant room for improvement in personalized mobility applications.
We publicly release the benchmark data, evaluation toolkit, and documentation at \url{https://github.com/AMAP-ML/MobilityBench}.

\end{abstract}



\maketitle

\section{Introduction}


\begin{figure*}[t]
  \centering
  \includegraphics[width=0.9\textwidth,height=0.37\textheight,keepaspectratio]{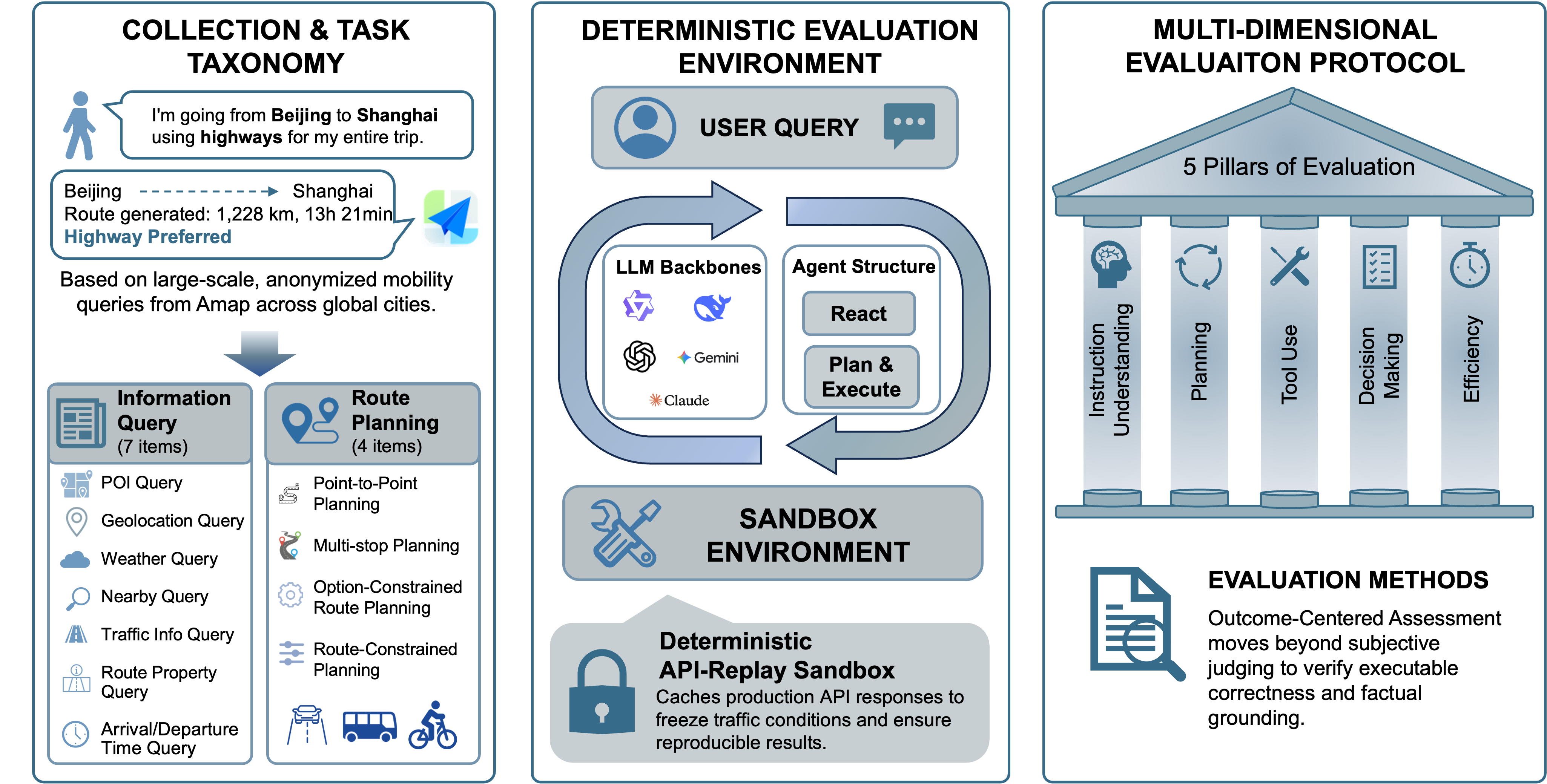}
  \caption{Overview of MobilityBench, a systematic benchmark for evaluating route-planning agents.}
  \label{fig:main}
\end{figure*}

The advance of large language models (LLMs) has catalyzed the emergence of tool-augmented agents, which integrate natural language reasoning with executable actions via external APIs~\cite{schick2023toolformer, patil2024gorilla}. By grounding user intent in programmatic interactions with real-world services, such agents substantially broaden the range of tasks they can support, from simple information retrieval to complex decision-making workflows, such as web navigation~\cite{ma2023laser, shen2025workflowagent}, computer interaction~\cite{hu2024dawn, lu2025axis}, and route planning~\cite{chen2024travelagent, zhe2025constraint}.

Among these agents, route-planning agents constitute a particularly challenging application domain, operating under diverse and dynamic real-world constraints that shape everyday human mobility~\cite{xie2024travelplanner, chaudhuri2025tripcraft, cheng2025travelbench}. Real-world mobility requests extend far beyond simple point-to-point navigation~\cite{yu2025dsfnet}, often involving multiple, interacting constraints, such as user preferences (e.g., avoiding highways or minimizing transfers), ordered waypoints, modality-dependent conditions, and time-sensitive requirements. Addressing such demands requires agents to accurately interpret nuanced user instructions, invoke appropriate travel-related APIs, and generate executable itineraries with reliable cost estimates—including travel time, distance, and transfer counts—capabilities that remain difficult to evaluate systematically in realistic mobility settings.

Recent benchmarks for evaluating the planning capabilities of LLMs and agents, such as TravelBench~\cite{cheng2025travelbench} and TravelPlanner~\cite{xie2024travelplanner}, primarily focus on high-level itinerary generation and abstract constraint reasoning. As a result, they fall short of capturing the complexity of route planning for everyday human mobility, which requires fine-grained reasoning over large-scale, map-based environments and dynamically changing conditions. Meanwhile, systematically evaluating route-planning agents in real-world mobility scenarios still faces several fundamental challenges: (1)~\textbf{scalable scenario coverage}, as evaluation must span route-planning problems of varying difficulty and combinations of constraints, ranging from simple point-to-point queries to complex multi-constraint requests; (2)~\textbf{non-determinism of live mapping APIs}, whose responses vary over time due to traffic dynamics, service availability, and backend updates~\cite{yao2022webshop, liu2023agentbench}, thereby undermining reproducibility and fair comparison; (3)~\textbf{comprehensive and reliable evaluation}, as effective assessment requires integrating multiple objective criteria beyond LLM-based subjective judging~\cite{zheng2023judging} to verify API-call validity, constraint satisfaction, and factual grounding; and (4)~\textbf{extensible and reproducible evaluation toolkit}, as rapid advances in LLM backbones and agent frameworks demand a lightweight, modular toolkit that supports easy deployment, scalable data expansion, and consistent evaluation across settings.

To address these challenges, we introduce \textbf{\ModelName}, a scalable benchmark for evaluating route-planning agents in real-world mobility scenarios. \ModelName\ is constructed from large-scale, anonymized real user queries collected from Amap, one of the largest map and navigation service providers in China, and is designed to reflect the diversity and complexity of everyday mobility needs while removing all personally identifiable information. It covers a broad spectrum of real-world route-planning intents, including point-to-point routing, customized multi-waypoint itineraries, and multimodal route planning that integrates driving, walking, cycling, and public transit. In addition, \ModelName\ supports preference-aware navigation, such as avoiding highways or minimizing transfers, as well as mobility-related information access, including bus station details, bus line information, and road congestion status. The benchmark spans queries from over 350 cities worldwide and is designed to be easily extensible, enabling continuous expansion to new regions, scenarios, and intent types.

Given the inherent non-determinism and reproducibility challenges of live mapping services, \ModelName\ is built around a \textsl{deterministic API-replay sandbox} that enables reproducible, end-to-end evaluation of route-planning agents. During dataset construction, responses fromrouting and points-of-interest APIs are captured and cached through a standardized interface, effectively freezing traffic conditions and service states at the time of collection. During evaluation, all API calls issued by an agent are intercepted and resolved against the cached response store, ensuring that identical inputs consistently yield identical, verifiable outputs. By eliminating uncontrolled environmental variance introduced by live services, this sandbox-based design ensures that measured performance faithfully reflects an agent’s reasoning and tool-use capabilities rather than fluctuations in external systems.

We further propose a multi-dimensional evaluation protocol that centers on \textsl{outcome validity} while providing complementary assessments of \textsl{instruction understanding}, \textsl{planning}, \textsl{tool use}, and \textsl{efficiency}. This protocol integrates multiple objective criteria to verify executable correctness, constraint satisfaction, and grounded API usage, enabling fine-grained and reliable assessment beyond surface-level plausibility. To facilitate reproducible research and rapid iteration, we publicly release the benchmark data, evaluation toolkit, and documentation at \url{https://github.com/AMAP-ML/MobilityBench}, supporting easy deployment, extensibility to new agent frameworks, and consistent comparison across models and settings.

\section{Related Work}
\subsection{Route Planning in Urban Computing}
Route planning is a long-standing problem in urban computing, attracting sustained attention from both academia and industry due to its central role in large-scale transportation systems and location-based services. Early studies primarily focused on optimizing physical costs, such as distance or travel time, under a graph-theoretic setting. Classical shortest-path algorithms, including Dijkstra~\cite{dljkstra1959note} and A*~\cite{hart1968formal,delling2009engineering}, were widely adopted to guarantee optimality while improving scalability in real-world road networks. These methods established the algorithmic foundations of modern navigation systems, but typically assume homogeneous objectives and well-defined cost functions. As mobility demands became increasingly diverse, subsequent research moved beyond single-objective optimization toward preference-aware route planning. These approaches incorporate user interests and contextual factors by integrating routing with recommendation models, such as INTSR~\cite{yan2025intsr}. Nevertheless, most existing methods rely on structured features or predefined preference spaces, which limit their ability to accommodate long-tail, ambiguous, or weakly specified requirements expressed in natural language. 
Recently, LLMs have been explored as a new interface for route planning, owing to their strong capability in understanding complex semantic instructions. However, prior work has shown that LLMs alone are unreliable for spatial reasoning and constrained optimization in geographic settings~\cite{huang2024can,agrawal2025can}. To mitigate these limitations, hybrid frameworks have been proposed that couple LLMs with traditional planners, using LLMs for high-level decision guidance~\cite{meng2024llm,zeng20251000} or intent and constraint extraction~\cite{yuan2025llmap}. Further studies introduce hierarchical planning architectures~\cite{zhe2025constraint} and reinforcement learning–based optimization strategies~\cite{qu2025tripscore,ning2025deeptravel,chu2026gpg,dai2026harder} to improve robustness under multiple objectives and constraints. In parallel, tool-augmented language agents have demonstrated strong capabilities in interacting with real-world systems and coordinating external tools for structured decision-making, making them a promising paradigm for route planning in real-world mobility scenarios. Existing travel planning agents, however, mainly focus on high-level itinerary generation and abstract constraint reasoning, without tightly integrating semantic intent understanding with low-level route optimization over real road networks. As a result, they fall short of capturing the complexity of route planning required for everyday human mobility. In this work, we introduce MobilityBench, a scalable benchmark for evaluating LLM-based route-planning agents in real-world mobility scenarios, to advance research in this area.

\subsection{Tool-augmented Agent Benchmark}
Building on the emergence of tool-augmented agents enabled by LLMs, recent work has focused on evaluating agents’ ability to follow instructions and interact with external tools. For instance, ToolBench~\cite{qin2023toolllm} constructs a large-scale benchmark over real-world APIs, requiring agents to perform sequential search and planning to complete complex instructions. $\tau$-bench~\cite{yao2024tau} further emphasizes interactive evaluation by simulating user–agent interactions and measuring behavioral consistency across repeated trials. In contrast to these general-purpose evaluations, recent work in urban computing has proposed domain-specific benchmarks for agent evaluation. TravelPlanner introduces a benchmark for multi-day itinerary construction by integrating domain-specific tools such as flight and restaurant search, and evaluates agents under itinerary-level environmental, commonsense, and hard constraints~\cite{xie2024travelplanner}. TravelBench further extends this task to multi-turn dialogue scenarios, enabling the evaluation of agents’ ability to infer and refine users’ implicit preferences through interaction~\cite{cheng2025travelbench}. Despite these advances, existing benchmarks primarily focus on high-level itinerary generation and abstract constraint satisfaction, and do not systematically evaluate agents’ ability to perform fine-grained route planning under mobility-specific constraints, such as preference-aware routing (e.g., avoiding highways or minimizing transfers), ordered waypoint requirements, modality-dependent conditions, and time-sensitive constraints. To address this gap, we introduce MobilityBench, a scalable benchmark designed to evaluate LLM-based route-planning agents in real-world mobility scenarios.

\definecolor{rowgreen}{HTML}{EAF8E9}
\definecolor{rowyellow}{HTML}{FFF9E6}

\begin{table*}[t]
  \centering
  \caption{Overview of task scenarios in \ModelName, grouped by intent family.}
  \label{tab:gis_scenarios}
  \small
  \setlength{\tabcolsep}{6pt}
  \renewcommand{\arraystretch}{1.15}
  \newcommand{\shade}{\cellcolor{gray!8}}
  \begin{tabularx}{\textwidth}{@{} p{3.1cm} l X @{}}
    \toprule
    \textbf{Intent Family} & \textbf{Task Scenario} & \textbf{Example Query} \\
    \midrule
    \multirow{5}{=}{\textbf{Basic Information\\Retrieval}}
      & POI Query              & Where is the gas station? \\
      & \shade Geolocation Query      & \shade Where am I? \\
      & Nearby Query           & Search for restaurants near Beijing Capital International Airport. \\
      & \shade Weather Query          & \shade What is the weather like in Wuhan tomorrow? \\
      & Traffic Info Query     & Is there a traffic jam on Chengdu Avenue right now? \\
    \midrule
    \multirow{2}{=}{\textbf{Route-Dependent\\Information Retrieval}}
      & \shade Route Property Query          & \shade How far is it from Hefei to Huangshan? \\
      & Arrival/Departure Time Query  & If I drive from my home to Capital International Airport now, when will I arrive? \\
    \midrule
    \multirow{2}{=}{\textbf{Basic Route Planning}}
      & \shade Point-to-Point Planning & \shade Drive from Tiananmen Square to Capital International Airport. \\
      & Multi-stop Planning     & Route planning starting from No.\ 60 Qinhe Road, via Yinji Mall and Zhenghong City. \\
    \midrule
    \multirow{2}{=}{\textbf{Preference-Constrained\\Route Planning}}
      & \shade Option-Constrained Route Planning & \shade Plan a driving route to Shanghai Disneyland that avoids tolls/highways. \\
      & Route-Constrained Planning        & Route to Shanghai Disneyland via People's Square, avoid Inner Ring Elevated Road. \\
    \bottomrule
  \end{tabularx}
\end{table*}
\section{\ModelName}
MobilityBench is a scalable benchmark for evaluating route-planning agents in real-world mobility scenarios. We first describe how the benchmark is built from large-scale anonymized mobility queries and organized into a comprehensive task taxonomy. We then introduce a structured ground-truth representation that explicitly captures the minimal tool interactions and intermediate evidence required to correctly resolve each request, serving as a stable and interpretable reference for evaluation. To ensure reproducibility, all tool interactions are executed within a deterministic replay sandbox. Finally, we present a multi-dimensional evaluation protocol that leverages this structured ground truth to assess agent performance.


\subsection{Benchmark Construction}
\subsubsection{Episode-centric Formulation.}
To enable rigorous evaluation of route-planning agents in realistic mobility scenarios, \ModelName\ adopts an episode-centric formulation, in which each episode encapsulates a self-contained mobility request solvable via tool augmentation. Formally, an episode is represented as a four-tuple $e = (x, z, \mathcal{S}, y)$, where:
\begin{itemize}[left=0pt]
    \item $x$ denotes an anonymized natural-language user query;
    \item $z$ encodes contextual information associated with the request, such as user location, city, and other background variables relevant to mobility decision-making;
    \item $\mathcal{S}$ denotes a fixed and replayable snapshot of relevant API responses provided by the replay sandbox (Section~\ref{seq:sandbox}), enabling consistent and deterministic evaluation across agent runs; and
    \item $y$ denotes a structured ground-truth annotation constructed (Section~\ref{sec:groundtruth}) and used exclusively to support the automated evaluation protocol (Section~\ref{sec:eval_protocal}). It is never exposed to the agent and serves solely for evaluation and diagnostic analysis.
\end{itemize}
Throughout this work, the route-planning agents are not permitted to ask users for clarification. Consequently, all episodes are designed to be fully solvable based solely on the initial user query $x$.


\subsubsection{Data Collection and Task Taxonomy Construction}
\label{sec:Intent_Taxonomy}
MobilityBench is constructed from large-scale, anonymized mobility queries collected from AMap over the past six months. In real-world on-the-go scenarios such as driving or walking, safety and convenience constraints limit users’ ability to interact with mobile devices, making voice a natural and prevalent input modality for expressing mobility intent. As a result, voice queries provide direct and largely unconstrained expressions of real user intent, encompassing destination goals, situational information needs, and explicit preference constraints.
In our dataset, these voice queries are transcribed into text and treated uniformly as query inputs for subsequent processing.
From a large corpus of raw queries, we construct the benchmark through a multi-stage filtering and curation pipeline, resulting in a substantial collection of high-quality episodes.
Under a strict no-clarification assumption, where each query must be self-contained and solvable without follow-up interaction, we remove malformed, underspecified, or ambiguous requests and deduplicate near-identical queries to ensure diversity.

Following this approach, we leverage Qwen-4B to perform intent classification over the curated queries, identifying diverse real-world mobility scenarios that define the task taxonomy of our benchmark. Specifically, we initialize the process with two coarse-grained intent roots: information access (e.g., POI, traffic, and weather lookup) and route planning (e.g., navigation to a destination). To identify long-tail and previously unobserved intents, we adopt an open-set labeling protocol, whereby queries that cannot be aligned with existing labels prompt the model to propose new candidate intents along with concise definitions. These candidate labels are subsequently iteratively consolidated, merged, and refined through multiple rounds of expert adjudication, ensuring semantic clarity, mutual exclusivity, and comprehensive coverage of the intent space. The resulting taxonomy comprises 11 task scenarios, which are further organized into four high-level Task families:

\begin{itemize}[left=0pt]
\item \textbf{Basic Information Retrieval}, which encompasses fundamental information-seeking tasks, including \textsl{POI Query}, \textsl{Geolocation Query}, \textsl{Nearby Query}, \textsl{Weather Query}, and \textsl{Traffic Info Query}.
\item \textbf{Route-Dependent Information Retrieval}, which targets information needs that require computing a route as an intermediate step, including \textsl{Route Property Query} (e.g., distance or path characteristics) and \textsl{Arrival/Departure Time Query}.
\item \textbf{Basic Route Planning}, which consists of two standard navigation tasks: \textsl{Point-to-Point Planning}, routing from a single origin to a single destination, and \textsl{Multi-stop Planning}, routing across multiple intermediate destinations.
\item \textbf{Preference-Constrained Route Planning}, which covers route planning tasks involving explicit user-specified preferences or constraints beyond basic navigation. This family includes \textsl{Option-Constrained Route Planning}, which applies tool-native, standardized routing options such as minimizing tolls, preferring highways, optimizing for the fastest route, fewer transfers, or less walking; and \textsl{Route-Constrained Planning}, which enforces explicit path-level constraints specified by users, such as required waypoints or excluded roads.
\end{itemize}
Table~\ref{tab:gis_scenarios} presents representative examples for each task scenario, while detailed scenario definitions and additional examples are provided in Appendix~A, Table~\ref{tab:all_scenarios}.

\subsubsection{Ground-Truth Construction.}
\label{sec:groundtruth}
To enable automated evaluation, we construct a structured ground-truth annotation $y$ for each episode following scenario-specific standard operating procedures (SOPs) defined by domain experts, which specify the minimal sequence of tool interactions required to correctly resolve a query. 
Specifically, we construct a scenario-specific standard tool program that defines the minimal sequence of tool calls required to answer a query. The workflow operationalizes the corresponding SOP as a structured and executable program, executes it within an existing agent framework to orchestrate tool invocations, validates the resulting outputs against historical data with reliability filtering, and consolidates the full execution trace together with key intermediate artifacts into a ground-truth archive. The standard tool program consists of three core steps: (i) extracting and normalizing query slots such as points of interest, temporal constraints, travel modes, and user preferences; (ii) resolving textual locations into structured entities or geographic coordinates via POI retrieval or geocoding tools; and (iii) after parameter validation, invoking downstream tools including routing, real-time traffic, and weather services while verifying constraint feasibility when applicable. The resulting tool evidence is then converted into a structured reference $y$ for automated evaluation and diagnostic analysis.

\subsubsection{Deterministic Replay Sandbox.}
\label{seq:sandbox}

During ground-truth construction, we rely on tools provided by the AMap Web Service API\footnote{\url{https://lbs.amap.com/api/webservice/summary}} to derive reference outputs. During evaluation, however, agents are prohibited from querying live API endpoints, as real-time updates (e.g., dynamic traffic and weather conditions) and external factors (e.g., API rate limits) would otherwise introduce non-determinism and compromise fair and reproducible comparisons. Instead, all tool interactions are routed through a deterministic replay sandbox that serves pre-recorded, contextually consistent responses.

The replay sandbox returns responses captured during ground-truth execution and ensures deterministic behavior across agent runs. Each tool invocation is resolved from a pre-recorded cache keyed by canonicalized arguments, such as normalized coordinates and standard time formats. When an exact cache hit is unavailable, the sandbox applies task-appropriate fallback strategies, including fuzzy matching for entity-based queries and nearest-neighbor spatial matching for coordinate-based queries, subject to a maximum distance threshold. All tool invocations undergo strict schema validation, including required-field checks and type and range constraints. Calls that fail validation or cannot be resolved are treated as tool-use failures and are explicitly reflected in the evaluation metrics (Section~\ref{sec:eval_protocal}), enabling fair and reproducible evaluation.
\subsubsection{Dataset Statistics.}
After constructing ground-truth, we filter out episodes whose answers cannot be reliably obtained or verified via tool execution, retaining only episodes with executable and checkable outcomes. As a result, \ModelName\ contains 100,000 episodes covering diverse geographic regions. Specifically, our benchmark spans \textbf{22} countries and over \textbf{350} cities (including metropolitan areas), with a long-tailed distribution across locations. We report the scenario distribution across the 11 intents
, where \textbf{36.6\%} of episodes belong to Basic Information Retrieval tasks, \textbf{9.6\%} to Route-Dependent Information Retrieval tasks, \textbf{42.5\%} to Basic Route Planning tasks, and \textbf{11.3\%} to Preference-Constrained Route Planning tasks.



\begin{figure}
    \centering
    \vspace{-5mm}
    \includegraphics[width=\linewidth]{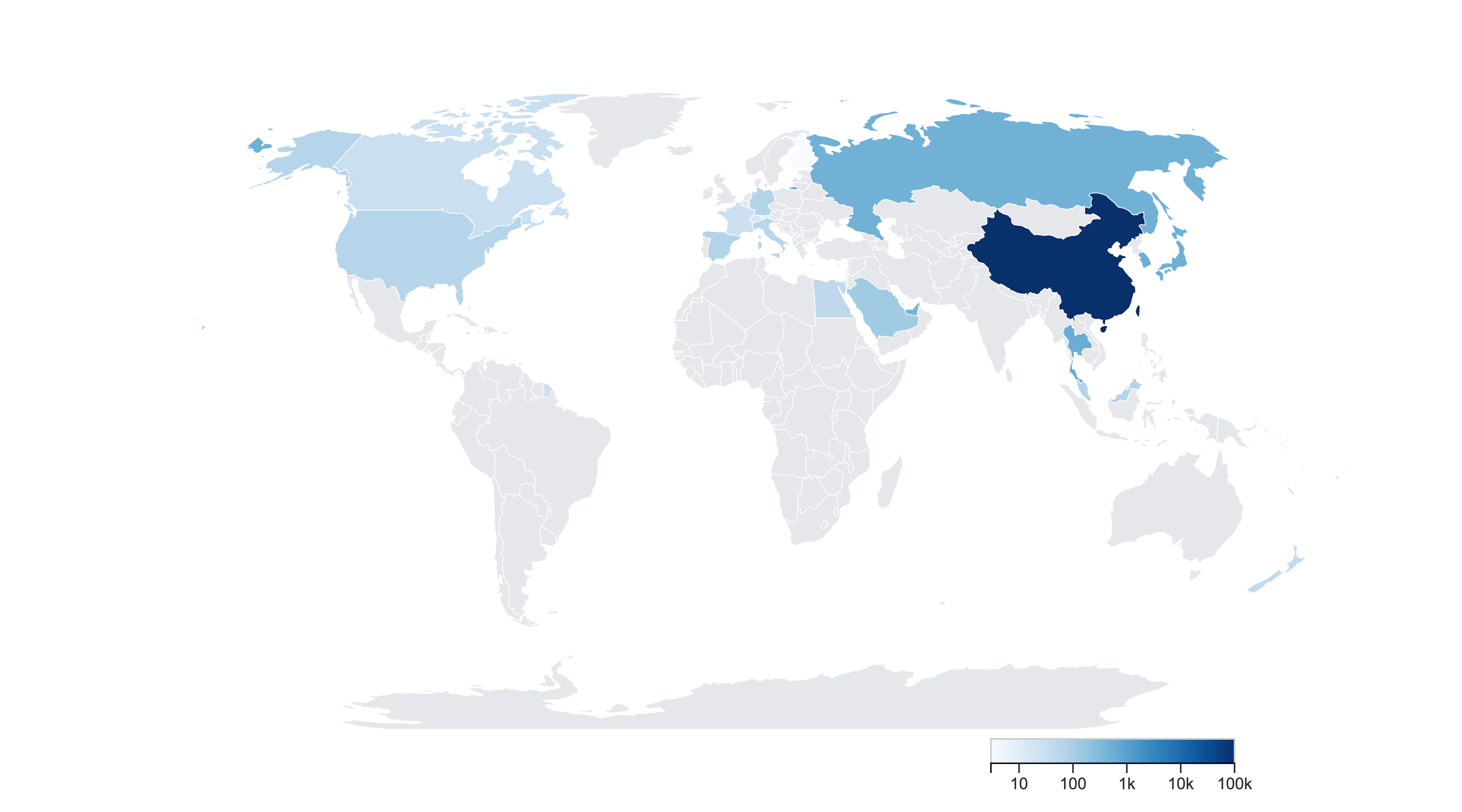}
    \vspace{-5mm}
    \caption{Global coverage of MobilityBench Data.}
    \vspace{-2mm}
    \label{fig:heatmap_world}
\end{figure}


\subsection{Evaluation Protocol}
\label{sec:eval_protocal}
To enable a comprehensive and in-depth evaluation of route-planning agents across diverse mobility scenarios, we introduce a multi-dimensional evaluation protocol. Existing evaluations predominantly rely on end-to-end success rates, which treat agent behavior as a black box and obscure the intermediate failures along the decision-making chain. Such coarse-grained metrics are insufficient for diagnosing the complex reasoning processes required in realistic route planning tasks. To address this limitation, our protocol decomposes an agent’s behavior into four core capabilities: \textsl{Instruction Understanding}, \textsl{Planning}, \textsl{Tool Use}, and \textsl{Decision Making}, corresponding to the key stages of route-planning reasoning. Each capability is further quantified using a set of fine-grained indicators, enabling precise diagnosis of performance bottlenecks and failure modes that are invisible to end-to-end metrics.

\subsubsection{Instruction Understanding}
Since accurate interpretation of user requirements is a prerequisite for route planning, we first evaluate the agent’s instruction understanding capability. Drawing on standard paradigms in natural language understanding~\cite{jbene2025intent,icmlt2020proceedings}, this capability is assessed through two indicators, detailed as follows:


\noindent\textbf{Intent Detection (ID).} We quantify the agent’s ability to understand the instructional intent embedded in a user query. Specifically, the agent is explicitly instructed to output a set of intent labels corresponding to the task scenario categories defined in Section~\ref{sec:Intent_Taxonomy}. We measure intent detection by comparing the agent’s predicted intent label $\hat{y}_{\mathrm{ID}}(x)$ with the ground-truth intent label $y_{\mathrm{ID}}(x)$ for each query $x$. A prediction is considered correct if the similarity between the two labels exceeds a predefined threshold $\alpha_{\mathrm{threshold}}$. The overall intent detection score is computed as:
\begin{equation}
\mathrm{ID} = \frac{1}{|\mathcal{X}|} \sum_{x \in \mathcal{X}}
\mathbb{I}\!\left( \mathrm{sim}\!\left(\hat{y}_{\mathrm{ID}}(x),\, y_{\mathrm{ID}}(x)\right)
\ge \alpha_{\mathrm{threshold}} \right).
\end{equation}


\noindent\textbf{{Information Extraction (IE)}.} This indicator evaluates an agent’s ability to extract explicit and implicit constraints from user queries, including spatial attributes (e.g., origins and destinations), temporal parameters (e.g., departure windows and duration constraints), and preference-related signals (e.g., traffic avoidance or modality priorities). For a query $x$, let $\hat{y}_{\mathrm{IE}}(x)$ and ${y}_{\mathrm{IE}}(x)$ denote the predicted and ground-truth constraint sets, respectively. An extraction is considered correct only if the two sets exactly match. The overall IE score is computed as:
\begin{equation}
\mathrm{IE} = \frac{1}{|\mathcal{X}|} \sum_{x \in \mathcal{X}}
\mathbb{I}\!\left(\hat{y}_{\mathrm{IE}}(x) = y_{\mathrm{IE}}(x)\right).
\end{equation}


\subsubsection{Planning}
Effective planning is a core capability of LLM-based agents, especially in real-world mobility scenarios where route planning requires multi-step reasoning under uncertainty. This dimension evaluates the agent’s ability to generate a logically coherent and sequential execution plan for complex routing tasks.


\noindent\textbf{Task Decomposition (DEC).} 
This dimension evaluates an agent’s ability to decompose a high-level user goal into a coherent sequence of atomic actions, reflecting whether the agent produces the right steps without omissions or redundancy. Given a predicted action sequence $V_{pred}(x) = \{v_1, v_2, ..., v_n\}$ and the corresponding ground-truth sequence $V_{gold}(x)$, we assess task decomposition quality by jointly considering step coverage and step correctness, that is,
\begin{equation}
\begin{split}
 \text{DEC-P} = \frac{1}{|\mathcal{X}|} \sum_{x \in \mathcal{X}} \frac{|V_{pred}(x) \cap_{f_{DEC}} V_{gold}(x)|}{|V_{pred}|}, \\
 \text{DEC-R} = \frac{1}{|\mathcal{X}|} \sum_{x \in \mathcal{X}} \frac{|V_{gold}(x) \cap_{f_{DEC}} V_{pred}(x)|}{|V_{gold}|}, \\
\end{split}
\end{equation}
where $A\cap_{f_{DEC}}B = \{a \in A \mid \exists b \in B, f_{DEC}(a,b)=\mathrm{True} \} $ and $f_{DEC}(\cdot,\cdot)$ is a function that determines whether two atomic actions are considered a match.

\subsubsection{Tool Use}
Tool invocation serves as the interface between the agent and the sandbox environment. To comprehensively evaluate an agent’s tool invocation capability, we define three evaluation indicators: tool selection, schema compliance, and parameter filling.

\noindent\textbf{Tool Selection (TS).}
This metric evaluates whether an agent correctly identifies the required tool(s) from a candidate tool set $\mathcal{T}$ based on the inferred user intent. Let $T_{pred}(x)$ denote the set of tools selected by the agent, and $T_{gold}(x)$ denote the ground-truth set of required tools. We measure tool selection quality from two complementary aspects: coverage and redundancy. Coverage reflects whether all necessary tools are selected, while redundancy penalizes unnecessary tool calls, (for easier comparison, we report redundancy as its complement. 
\begin{equation}
\begin{split}
 \text{TS-P} = \frac{1}{|\mathcal{X}|} \sum_{x \in \mathcal{X}} \frac{|T_{pred}(x) \cap T_{gold}(x)|}{|T_{pred}|}. \\
 \text{TS-R} = \frac{1}{|\mathcal{X}|} \sum_{x \in \mathcal{X}} \frac{|T_{gold}(x) \cap T_{pred}(x)|}{|T_{gold}|}, \\
\end{split}
\end{equation}

\noindent\textbf{Schema Compliance (SC).} This metric evaluates whether an agent’s tool invocation conforms to predefined API specifications, requiring that all mandatory parameters are provided and that their values fall within valid formats and ranges. For each query $x$, let $ST_{pred}(x)$ denote the sequence of tool invocations produced by the agent, and let $P(t)$ denote the set of parameters associated with each tool call $t \in ST_{pred}(x)$. We define $f_{{SC}}(P(t),t)$ as an indicator function that determines whether the parameters provided for a tool $t$ conform to the predefined valid formats and ranges. Along this line, the overall SC score is calculated by:
\begin{equation}
\mathrm{SC}
= \frac{1}{|\mathcal{X}|}
\sum_{x \in \mathcal{X}}
\frac{1}{\left| ST_{pred}(x) \right|}
\sum_{t \in ST_{pred}(x)}
f_{{SC}}\!\left(P(t),\, t\right).
\end{equation}

\subsubsection{Decision Making}
Decision quality evaluates whether an agent can produce a final solution and whether that solution is correct. We assess this dimension using the following two metrics:

\noindent\textbf{Delivery Rate (DR).} This indicator measures the proportion of queries for which an agent successfully generates a complete and executable final output (e.g., a full itinerary) without interruption or tool invocation failure. This metric reflects the agent’s ability to complete the end-to-end task pipeline.

\noindent\textbf{Final Pass Rate (FPR).} This indicator evaluates the effectiveness of the generated solution. A solution is considered successful only if it satisfies all user-specified explicit and implicit constraints, capturing the agent’s ability to produce a valid final outcome.

\subsubsection{Efficiency}
In addition to agent behavioral quality, we evaluate efficiency to characterize computational overhead and practical deployability. We consider the following indicator:

\noindent\textbf{Input Token (IT).} 
This metric measures the cumulative volume of contextual information processed by the model, including system prompts, task instructions, and the historical trajectory of observations and actions. A higher IT count typically reflects a heavier reliance on long-context reasoning or a more verbose feedback loop.
\noindent\textbf{Output Token (OT).}
This metric quantifies the total number of tokens generated by the model. While higher OT indicate more thorough reasoning, it also implies increased generation time and resource consumption. 



\section{Experiments}

\definecolor{colorOpen}{RGB}{252, 232, 210}
\definecolor{colorClosed}{RGB}{220, 225, 255}
\definecolor{colorAgent}{RGB}{220, 250, 220}
\definecolor{colorSFT}{RGB}{250, 210, 215}

\def\rID{88.70} \def\rIE{93.06} \def\rDP{84.12} \def\rDR{81.56} \def\rTP{90.74} \def\rTR{84.66} \def\rSC{98.70} \def\rD{85.95} \def\rFP{69.09} \def\rIT{15013.79} \def\rOT{560.19}

\def\pID{97.28} \def\pIE{96.58} \def\pDP{89.62} \def\pDR{74.68} \def\pTP{85.44} \def\pTR{78.17} \def\pSC{97.86} \def\pD{83.53} \def\pFP{65.77} \def\pIT{12394.29} \def\pOT{667.69}

\begin{table*}[t]
\centering
\small
\setlength{\tabcolsep}{5pt} 
\caption{Performance of models on MobilityBench.  Abbreviations: Instr. Und. for Instruction Understanding; Dec. Mak. for Decision Making; ID for Intent Detection; IE for Information Extraction; DEC for Task Decomposition; TS for Tool Selection; SC for Schema Compliance; DR for Delivery Rate; FPR for Final Pass Rate; IT for Input Token; and OT for Output Token.}
\label{tab:main_results}

\begin{tabular}{lccccccccccc} 
\toprule
 & \multicolumn{2}{c}{\textbf{Instr. Und.}}
 & \multicolumn{2}{c}{\textbf{Planning}} 
 & \multicolumn{3}{c}{\textbf{Tool Use}} 
 & \multicolumn{2}{c}{\textbf{Dec. Mak.}}
 & \multicolumn{2}{c}{\textbf{Efficiency}}\\ 
\cmidrule(lr){2-3} \cmidrule(lr){4-5} \cmidrule(lr){6-8} \cmidrule(lr){9-10}\cmidrule(lr){11-12}

\textit{Model} & \textit{ID} & \textit{IE} & \textit{DEC-P} & \textit{DEC-R} & \textit{TS-P} & \textit{TS-R} & \textit{SC} & \textit{DR} & \textit{FPR} & \textit{IT} & \textit{OT} \\ \midrule


\rowcolor{colorOpen} \multicolumn{12}{c}{\textit{ReAct}} \\
\begin{filecontents*}{table/main_format.csv}
Model,Group,ID,IE,DEC-P,DEC-R,TS-P,TS-R,SC,DR,FPR,IT,OT
GPT-4.1,React,85.86 ,90.07 ,75.53 ,74.14 ,82.38 ,81.92 ,97.00 ,79.23 ,61.66 ,18680.81 ,1166.27 
GPT-5.2,React,82.16 ,89.65 ,81.24 ,62.22 ,82.42 ,76.20 ,95.49 ,79.09 ,61.90 ,18304.90 ,1166.12 
Claude-Sonnet-4.5,React,88.70 ,93.06 ,80.71 ,74.76 ,82.83 ,82.99 ,97.42 ,80.62 ,63.17 ,18856.68 ,1311.01 
Claude-Opus-4.5,React,85.99 ,91.23 ,84.12 ,70.15 ,83.21 ,83.73 ,97.52 ,80.20 ,62.22 ,19672.63 ,1305.40 
Gemini-3-Flash-Preview,React,84.00 ,88.16 ,71.95 ,68.34 ,90.37 ,76.46 ,98.31 ,85.18 ,67.90 ,21072.79 ,1232.76 
Gemini-3-Pro-Preview,React,83.54 ,88.75 ,68.70 ,65.11 ,90.74 ,75.04 ,98.70 ,84.38 ,69.09 ,20164.76 ,1242.48 
DeepSeek-V3.2-Exp,React,78.18 ,90.78 ,71.85 ,77.19 ,87.99 ,82.19 ,98.23 ,84.95 ,68.88 ,15427.89 ,622.05 
Qwen3-4B,React,77.89 ,86.75 ,47.24 ,81.56 ,80.74 ,72.82 ,94.46 ,63.80 ,53.80 ,26078.99 ,657.78 
Qwen3-30B-A3B,React,74.73 ,91.23 ,70.60 ,72.93 ,84.04 ,83.35 ,97.06 ,84.57 ,66.65 ,15013.79 ,560.19 
Qwen3-32B,React,80.87 ,88.46 ,68.37 ,77.58 ,83.08 ,83.94 ,96.76 ,83.16 ,65.68 ,15544.50 ,583.22 
Qwen3-235B-A22B,React,82.13 ,90.51 ,72.23 ,77.75 ,84.13 ,84.66 ,97.24 ,85.95 ,66.69 ,15391.23 ,604.73 
GPT-4.1,PlanAndExecute,94.40 ,94.79 ,89.46 ,68.85 ,84.61 ,73.26 ,97.02 ,80.70 ,63.40 ,13426.36 ,747.35 
GPT-5.2,PlanAndExecute,89.58 ,96.58 ,81.90 ,74.68 ,81.12 ,75.94 ,95.94 ,77.26 ,59.81 ,15312.45 ,1644.18 
Claude-Sonnet-4.5,PlanAndExecute,97.21 ,95.69 ,89.46 ,71.81 ,84.63 ,78.17 ,96.89 ,81.96 ,64.31 ,13267.99 ,863.81 
Claude-Opus-4.5,PlanAndExecute,76.82 ,95.81 ,88.80 ,70.99 ,84.76 ,76.53 ,97.22 ,83.53 ,65.77 ,12643.41 ,808.83 
Gemini-3-Flash-Preview,PlanAndExecute,97.28 ,94.41 ,89.60 ,66.18 ,85.44 ,68.13 ,97.86 ,80.50 ,62.87 ,14515.42 ,784.06 
Gemini-3-Pro-Preview,PlanAndExecute,96.35 ,95.32 ,88.97 ,65.71 ,85.12 ,64.38 ,97.35 ,78.64 ,62.80 ,15936.49 ,815.26 
DeepSeek-V3.2-Exp,PlanAndExecute,96.93 ,95.92 ,89.62 ,69.55 ,83.83 ,75.28 ,97.23 ,80.73 ,63.06 ,12394.29 ,706.14 
Qwen3-4B,PlanAndExecute,95.98 ,94.53 ,86.83 ,73.26 ,81.64 ,69.20 ,96.88 ,78.06 ,59.55 ,13612.71 ,673.03 
Qwen3-30B-A3B,PlanAndExecute,95.56 ,94.35 ,83.91 ,71.19 ,82.91 ,68.97 ,97.48 ,78.81 ,60.60 ,14820.45 ,667.69 
Qwen3-32B,PlanAndExecute,96.03 ,94.63 ,86.83 ,66.98 ,84.25 ,69.80 ,97.17 ,80.24 ,62.43 ,13658.79 ,703.31 
Qwen3-235B-A22B,PlanAndExecute,97.24 ,94.36 ,89.39 ,66.96 ,84.59 ,73.49 ,97.01 ,81.22 ,64.16 ,12563.66 ,703.60 
\end{filecontents*}
\csvreader[
    column names={1=\ModelName, 2=\GroupName, 3=\vID, 4=\vIE, 5=\vDECP, 6=\vDECR, 7=\vTSP, 8=\vTSR, 9=\vSC, 10=\vDR, 11=\vFPR, 12=\vIT, 13=\vOT},
    filter strcmp={\GroupName}{React},
    late after line=\\ 
]{table/main_format.csv}{}{%
    \ModelName & \hmax{\vID}{\rID} & \hmax{\vIE}{\rIE} & \hmax{\vDECP}{\rDP} & \hmax{\vDECR}{\rDR} & \hmax{\vTSP}{\rTP} & \hmax{\vTSR}{\rTR} & \hmax{\vSC}{\rSC} & \hmax{\vDR}{\rD} & \hmax{\vFPR}{\rFP} & \hmin{\vIT}{\rIT} & \hmin{\vOT}{\rOT}
}

\rowcolor{colorClosed} \multicolumn{12}{c}{\textit{Plan and Execute}} \\
\csvreader[
    column names={1=\ModelName, 2=\GroupName, 3=\vID, 4=\vIE, 5=\vDECP, 6=\vDECR, 7=\vTSP, 8=\vTSR, 9=\vSC, 10=\vDR, 11=\vFPR, 12=\vIT, 13=\vOT},
    filter strcmp={\GroupName}{PlanAndExecute},
    late after line=\\
]{table/main_format.csv}{}{%
   \ModelName & \hmax{\vID}{\pID} & \hmax{\vIE}{\pIE} & \hmax{\vDECP}{\pDP} & \hmax{\vDECR}{\pDR} & \hmax{\vTSP}{\pTP} & \hmax{\vTSR}{\pTR} & \hmax{\vSC}{\pSC} & \hmax{\vDR}{\pD} & \hmax{\vFPR}{\pFP} & \hmin{\vIT}{\pIT} & \hmin{\vOT}{\pOT}
}


\bottomrule
\end{tabular}
\end{table*}

\subsection{Experimental Setup}
\subsubsection{Data Sampling}
Our benchmark is constructed from 100,000 episode $e$ collected from real-world mobility scenarios. To balance statistical significance with computational efficiency, we performed stratified random sampling across the 11 core performance analysis scenarios defined in Section~\ref{sec:Intent_Taxonomy}, while jointly enforcing stratification by city. Specifically, we strive to maintain a balanced sample distribution across scenarios while maintaining proportional coverage across diverse urban regions and city tiers, thereby mitigating geographic bias during scenario selection. This joint sampling strategy yields a final evaluation set of 7,098 episodes for subsequent analysis of agent performance.


\subsubsection{LLM Backbones}
We evaluated a diverse suite of representative \emph{open-source} and \emph{closed-source} LLMs as the backbones of rout-planning agents, spanning a broad range of model characteristics: (i) small- and large-parameter dense models, (ii) Mixture-of-Experts (MoE) architectures, and (iii) reasoning-oriented (\textit{Thinking}) models.

\noindent\textbf{Open-source backbones.} We evaluated the Qwen family (Qwen3-4B, Qwen3-30B-A3B, Qwen3-32B, Qwen3-235B-A22B) and DeepSeek models (DeepSeek-R1, DeepSeek-V3.2-Exp).

\noindent\textbf{Closed-source backbones.} We evaluated OpenAI GPT models (GPT-4.1, GPT-5.2), Anthropic Claude models (Claude-Opus-4.5, Claude-Sonnet-4.5), and Google Gemini models (Gemini-3-Pro-Preview, Gemini-3-Flash-Preview).
\subsubsection{Agent Implementations}
To evaluate the effectiveness of LLM-based agent workflows for route planning, we constructed route-planning agents based on two representative frameworks: ReAct~\cite{yao2022react} and Plan-and-Execute~\cite{wang2023plan}. At this stage, we did not incorporate alternative agent frameworks such as LLM Compiler, LATS, or Tree-of-Thought~\cite{kim2024llm,zhou2023language,yao2023tree,ji2026tree}. This design choice was motivated by two considerations. First, the selected frameworks are widely regarded as representative of mainstream agent reasoning pipelines, covering reactive and planning-based paradigms. Second, approaches such as LATS and Tree-of-Thought typically incur substantially higher computational overhead and exhibit limited adaptability to the task-specific constraints, tool interactions, and latency requirements inherent in our route-planning setting.


\subsubsection{Experimental Details}
To ensure reproducibility and fair comparison, we applied a unified set of evaluation settings across all LLM backbones and agent frameworks.

\noindent \textbf{Agent Inputs.} Each agent instance received the user query along with spatial context signals, such as city and geographic location. When tool use was enabled, we additionally provided structured tool schemas or invocation patterns to standardize tool usage across different frameworks and backbones.


\noindent \textbf{Model Configuration.} To further control evaluation variance, we set the sampling temperature to $0.1$ for all evaluated LLM backbones and capped the maximum output length at $8,192$ tokens.

\noindent \textbf{Agent Configuration.} To balance inference efficiency and robustness (e.g., preventing degenerate tool-calling loops), we limited the maximum number of inference steps to $10$. 

\begin{figure*}[t]
    \centering
    \includegraphics[width=\textwidth]{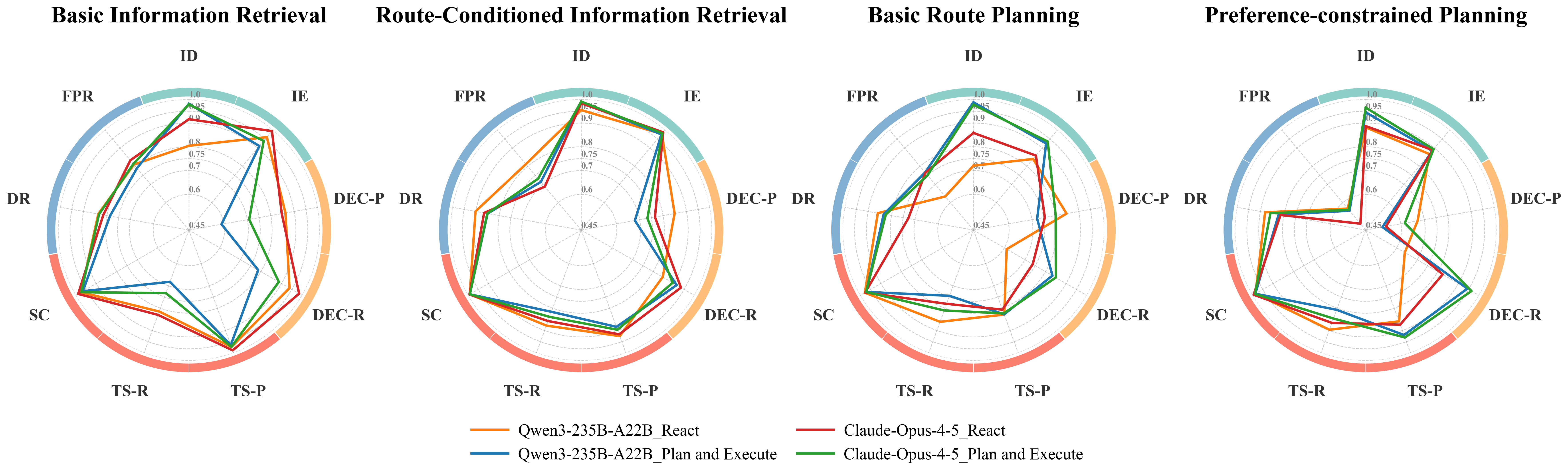}
    \vspace{-7mm}
    \caption{Performance across four high-level task families.}
    \vspace{-3mm}
    \label{fig:radars}
\end{figure*}

\subsection{Experimental Results}
\subsubsection{Overall Performance}\mbox{}\\[-\baselineskip]

\noindent\textbf{LLM performance.}
Under the Plan-and-Execute framework, Claude-Opus-4.5 stands out as the strongest performer, achieved a Delivery Rate of 83.53\% and a Final Pass Rate of 65.77\%, both the highest among all evaluated models in this setting. Within the ReAct framework, Gemini-3-Pro-Preview attained the highest FPR of 69.09\%. This result highlights its exceptional ability to preserve task-relevant context and maintain goal focus across extended iterative inference loops. 

\noindent\textbf{Closed-Source vs.\ Open-Source Models.}
As shown in Table \ref{tab:main_results}, Claude-Sonnet-4.5, Gemini-3-Pro-Preview still maintained a clear lead in instruction understanding dimensions, with average scores of 90.88\% and 88.61\% under the ReAct framework. However, the gap is narrowing significantly. Among open-source models, Qwen3-235B-A22B, a MoE architecture activating only 22B parameters per forward pass, achieved a DR of 85.95\% and an FPR of 66.69\% under the ReAct framework. Similarly, DeepSeek-V3.2-Exp demonstrated strong competitiveness, attaining an FPR of 68.88\% while maintaining substantially lower inference costs due to its efficient architecture. This provides a high-performance and cost-effective option for enterprise-level private deployments.

\noindent\textbf{Framework Comparison: ReAct vs.\ Plan-and-Execute.}
A systematic comparison of the two execution architectures reveals a fundamental trade-off between task success rate and computational efficiency. The final pass rate of the ReAct is generally better than that of Plan-and-Execute. This is mainly due to its closed-loop "think-act-observe" mechanism, which allows the agent to dynamically adjust its strategy based on real-time results returned by tools, while Plan-and-Execute's static pre-planning shows a significant lack of robustness when facing dynamic feedback in mobile scenarios. However, ReAct's superior robustness comes at a non-trivial computational cost.  Due to the continuous accumulation of observation history within the inference context, the average number of input tokens (IT) consumed by ReAct is significantly higher than that of Plan-and-Execute. Across all models, ReAct's average IT is approximately 35.38\% higher than Plan-and-Execute's. This increase translates directly into higher API costs and longer wall-clock inference times.

\subsubsection{Scenario Study}
To further reveal the capabilities of the model in different task scenarios, we created multi-dimensional indicator radar charts for four core categories in Figure \ref{fig:radars}, evaluating representative open-source and closed-source models under both ReAct and Plan-and-Execute frameworks.
The scene from left to right represents a significant increase in the depth of task logic and the complexity of constraints, and Preference-constrained Planning is the category where the model is the most likely to be error as we expected. In this type of tasks, Plan-and Execute framework performs best because it establishes a clear strategy in advance, which makes handling structured tasks with logical order more predictable and efficient, thereby suppressing illusions and trajectory deviations.



\subsubsection{Model Study}
We conduct a model-centric study to examine how model scaling and reasoning mode (Thinking vs.\ Non-thinking) influence route-planning agent performance on MobilityBench.
\noindent\textbf{Scaling effect.}
Experiments reveal a clear performance gap across model sizes (Table~\ref{tab:main_results}). Under the same dense architecture, scaling the base model from 4B to 32B yields a consistent improvement in average success rate, increasing by 0.91\%. Under the MoE setting, Qwen-30B-A3B further scales to Qwen-235B-A22B, bringing an additional gain of 5.43\% . Overall, these results align with the classic scaling law: increasing parameter scale leads to higher success rates in real-world mobility scenarios. 
By jointly examining DEC-P and DEC-R, we observe that, compared with smaller models, larger models tend to produce longer solution trajectories (i.e., more plans) to explore a broader space of possible outcomes. Although some of these steps can be redundant, this more exhaustive search-and-verification process ultimately improves the task success rate.

\begin{figure}
    \centering 
    \includegraphics[width=0.4\textwidth]{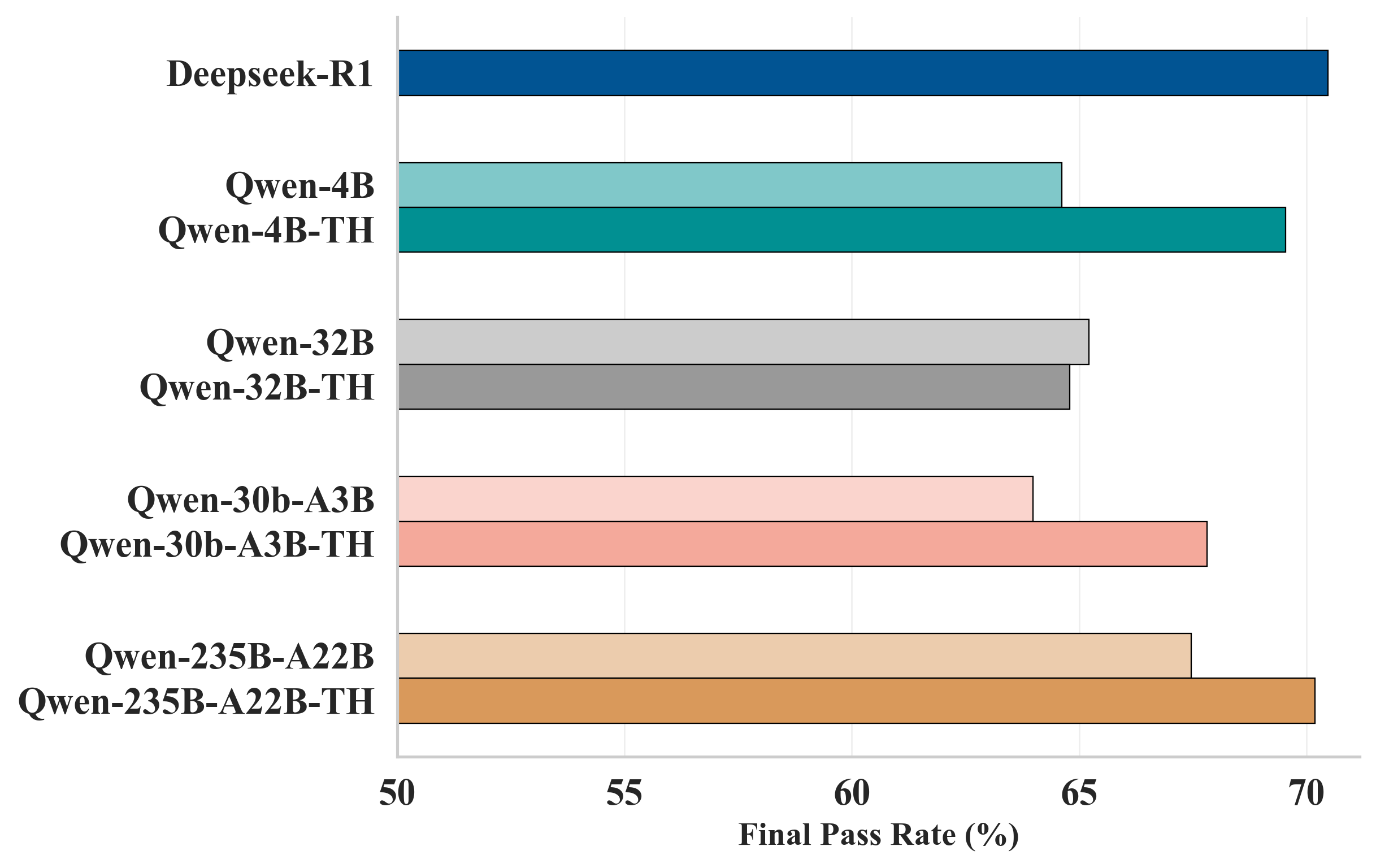}
    \vspace{-2mm}
    \caption{Final pass rate comparison (Thinking vs. Non-thinking) under the Plan-and-Execute framework.} 
    \vspace{-4mm}
    \label{fig:think} 
\end{figure}

\noindent\textbf{Thinking vs.\ Non-thinking.}
To examine the intrinsic potential of LLMs on complex route-planning tasks, we study the impact of reasoning mode (Thinking vs.\ Non-thinking) while accounting for the extra cost and latency introduced by Thinking. We sample 1,000 representative instances from MobilityBench for a controlled comparison, and evaluate how different reasoning patterns affect final task success. Figure~\ref{fig:think} reports the final pass rate of each model with and without Thinking enabled.

We evaluate Qwen-4B, Qwen-32B, Qwen-30B-A3B, and Qwen-235B-A22B under both settings, and additionally include DeepSeek-R1 as a strong reasoning-oriented baseline (Figure~\ref{fig:think}). DeepSeek-R1 achieves a final pass rate of 70.46\%, serving as a competitive reference point. Across models, enabling thinking consistently improves performance, with the largest gain observed for Qwen-30B-A3B, final pass rate increased by 5.98\% absolutely. 
Despite these gains, Thinking substantially increases the generated token volume, leading to markedly higher inference cost and latency. This overhead makes it challenging to deploy Thinking-enabled agents in real-time, production-grade online settings.

\section{Conclusion}
In this work, we presented MobilityBench, a scalable benchmark for the systematic evaluation of LLM-based route-planning agents in real-world mobility scenarios. Built from large-scale, anonymized real user queries, MobilityBench captured the diversity and complexity of everyday mobility demands while enabling reproducible, end-to-end evaluation through a deterministic API-replay sandbox. We further introduced a multi-dimensional evaluation protocol centered on outcome validity and complemented by assessments of instruction understanding, planning, tool use, and efficiency. Using MobilityBench, we evaluated multiple LLM-based route-planning agents across diverse real-world mobility scenarios and conducted an in-depth analysis of their behaviors and performance, revealing both their strengths and limitations under realistic conditions. MobilityBench provides a robust and extensible foundation for advancing research on route-planning agents and for enabling fair and reproducible comparison across LLMs and agent frameworks.

\clearpage
\bibliographystyle{ACM-Reference-Format}
\bibliography{main}

\appendix

\setcounter{table}{0}
\setcounter{figure}{0}
\renewcommand{\thetable}{S\arabic{table}}
\renewcommand{\thefigure}{S\arabic{figure}}



\section{Appendix}
\subsection{MobilityBench Task Scenarios}
To facilitate a thorough understanding of the benchmark's coverage and design rationale, we present a detailed taxonomy of task scenarios in Table~\ref{tab:all_scenarios}, including fine-grained subtypes and their definitions, and provide additional representative examples for each category, which are designed to reflect the diversity of natural language expressions that users may employ when issuing mobility-related instructions.

\newcolumntype{L}[1]{>{\raggedright\arraybackslash}p{#1}}

\newlist{titemize}{itemize}{1}
\setlist[titemize]{label=\textbullet, leftmargin=*, nosep, topsep=0pt, partopsep=0pt, itemsep=1pt}

\begin{table*}[t]
    \centering
    \small
    \caption{MobilityBench task scenarios. For each scenario, we provide a concise definition and representative user queries.}
    \label{tab:all_scenarios}

    \setlength{\tabcolsep}{8pt}
    \renewcommand{\arraystretch}{1.2}

    \rowcolors{2}{gray!6}{white}

    \begin{tabularx}{\textwidth}{@{\hspace{8pt}} L{0.17\textwidth} L{0.33\textwidth} X @{\hspace{8pt}}}

        \toprule
        \rowcolor{gray!12}
        \textbf{Scenario} & \textbf{Introduction} & \textbf{Query Examples} \\
        \midrule

        POI Search &
        Retrieve a point of interest (POI) by name or category and return key attributes (e.g., address, latitude/longitude). &
        \begin{titemize}
            \item Find a Starbucks.
            \item Search for a pharmacy in Nanshan District.
            \item Where is the shopping mall?
        \end{titemize} \\

        Geolocation Query &
        Reverse geocoding converts coordinates (or the current location) into an address, place name, and administrative region. &
        \begin{titemize}
            \item Give me my current location.
            \item Tell me where I am right now.
            \item What's the latitude and longitude of Beijing Railway Station?
        \end{titemize} \\

        Nearby Search &
        Find POIs within a specified radius of a target location. &
        \begin{titemize}
            \item Any parking lots within 500 meters of my location?
            \item Find the nearest EV charging station.
            \item Where is the nearest restroom nearby?
        \end{titemize} \\

        Weather Query &
        Query current weather and forecasts for a target area to support travel decisions. &
        \begin{titemize}
            \item I'm arriving in Hangzhou tomorrow—what's the weather like there?
            \item What's the temperature in Beijing tomorrow morning?
            \item Give a 3-day forecast for Shenzhen.
        \end{titemize} \\

        Traffic Info Query &
        Retrieve real-time traffic congestion information for roads or areas, including severity and affected segments. &
        \begin{titemize}
            \item How is traffic on Yan'an Elevated Road right now?
            \item Is there congestion near Guomao?
            \item How is the traffic flow on the way to the airport?
        \end{titemize} \\

        RouteProperty Query &
        Query attributes of a given route/itinerary (distance, duration, transfers, etc.). &
        \begin{titemize}
            \item How long from Lujiazui to Hongqiao by metro?
            \item How many transfers are there on this transit route?
            \item What's the distance to Jiuzhaigou Valley?
        \end{titemize} \\

        Arrival/Departure Time Query &
        Plan routes with time constraints (depart-at/arrive-by) and infer feasible schedules. &
        \begin{titemize}
            \item I must arrive at the airport by 7{:}30; when should I leave?
            \item My train departs at 9{:}00 PM tonight—what’s the best time to leave for Nanchang Railway Station?
            \item If I leave at 6 PM, can I reach the concert by 7?
        \end{titemize} \\

        Point-to-Point Planning &
        Plan a route from an origin to a destination under a specified travel mode. &
        \begin{titemize}
            \item How do I get from Pudong Airport to The Bund by subway?
            \item Drive from Tsinghua University to Sanlitun now.
            \item Bike from my location to Zhongshan Park.
        \end{titemize} \\

        Multi-stop Planning &
        Plan an ordered multi-stop route that visits multiple waypoints sequentially. &
        \begin{titemize}
            \item Start from the Grand Hyatt Beijing, stop at Wangfujing Department Store, then proceed to Beijing South Railway Station.
            \item Travel from Guangzhou South Railway Station to Chimelong Tourist Resort via Tianhe Sports Center.
        \end{titemize} \\

        Option-Constrained Route Planning &
        Plan routes based on standard user preferences supported by the routing API (e.g., avoid\_tolls, avoid\_highways, minimize\_transfers). &
        \begin{titemize}
            \item Drive to the zoo but avoid highways.
            \item Take public transit with at most one transfer.
            \item Find the cheapest route to the airport.
        \end{titemize} \\

        Customized Planning &
        Plan routes under bespoke constraints that must be satisfied (e.g., designated line/stop/segment). &
        \begin{titemize}
            \item I must take Metro Line 2; plan the route to the stadium.
            \item Route to the hospital via People's Square Station.
            \item Plan a route to the airport with the fewest traffic lights.
        \end{titemize} \\


        \bottomrule
    \end{tabularx}
\end{table*}

\subsection{Sandbox Tools}
A core design principle of MobilityBench is to evaluate agents within a realistic yet reproducible tool-use environment. To this end, we provide a comprehensive tool specification table as shown in Table~\ref{tab:map-tools-io}. It documents each tool used in the benchmark sandbox, including the tool name, input arguments and output fields. The sandbox tools are sourced from the AMap Open Platform. More detailed parameter definitions and response field descriptions are available in the official documentation\footnote{\url{https://lbs.amap.com/api/webservice/summary}}.

\begin{table*}[t]
\centering
\small
\caption{Overview of map-related tools and their tool-function I/O.}
\label{tab:map-tools-io}

\setlength{\tabcolsep}{8pt}
\renewcommand{\arraystretch}{1.2}

\begin{tabularx}{\textwidth}{L{0.16\textwidth} X X X}
\hline
\textbf{Tool} & \textbf{Function} & \textbf{Input} & \textbf{Output} \\ \hline

poi\_query & Search points of interest (POIs) using keywords, categories, city or city code. &
keyword(s), category, city, optional filters (e.g., limit). &
Candidate POIs: name, address, coordinates, category, brief metadata. \\ \hline

nearby\_poi\_query & Retrieve nearby POIs within a radius matching a category/keyword. &
center coordinate (lat/lon), radius, keyword/category, optional filters (e.g., limit/sort). &
Nearby POI list with distance (optional), name, address, coordinates, category. \\ \hline

reverse\_geocoding & Convert geographic coordinates into a human-readable address. &
coordinate (lat/lon). &
Address fields (province/city/district, street, number), nearby landmark/POI (optional), formatted address. \\ \hline

weather\_query & Query current weather or forecast for a location. &
city name or coordinate (lat/lon), time range/type (current/forecast). &
Weather report: temperature, precipitation, wind, humidity, conditions; air quality (optional). \\ \hline

traffic\_info\_query & Retrieve real-time/recent traffic conditions for a road segment/area. &
road segment/area identifier or polyline/bbox, optional time window. &
Traffic status: congestion level, speed, incidents/events (optional), timestamp, suggested impact on ETA (optional). \\ \hline

driving\_planning & Plan a driving route between origin and destination. &
origin (lat/lon), destination (lat/lon), optional waypoints, route preferences (avoid highways/tolls), traffic-aware flag. &
Route: distance, ETA, polyline/geometry, turn-by-turn steps, traffic-aware ETA (optional). \\ \hline

bus\_planning & Plan a public-transit route between origin and destination. &
origin (lat/lon), destination (lat/lon), departure time (optional), preferences (bus or subway, min transfers). &
Transit plan: lines, transfers, walking segments, total duration, fare/operating info (if available), step details. \\ \hline

bicycling\_planning & Plan a cycling route between origin and destination. &
origin (lat/lon), destination (lat/lon), optional preferences (bike lanes). &
Cycling route: distance, ETA, polyline/geometry, step-by-step directions, elevation/road-type hints (optional). \\ \hline

walking\_planning & Plan a walking route between origin and destination. &
origin (lat/lon), destination (lat/lon). &
Walking route: distance, ETA, polyline/geometry, step-by-step directions. \\ \hline

\end{tabularx}
\end{table*}



\end{document}